# PROOFING TOOLS TECHNOLOGY AT NEUROSOFT S.A.

**Christos Tsalidis**[*], **Giorgos Orphanos**[*], **Anna Iordanidou**[**] **and Aristides Vagelatos**[***]

[*] Neurosoft S.A.
24 Kofidou Str., 14231, N. Ionia, Athens, Greece
{tsalidis, orphan}@neurosoft.gr

[**] Department of Primary Education
University of Patras, 26500, Rion, Patras, Greece
A.Iordanidou@upatras.gr

[***] Research Academic Computer Technology Institute
13 Eptachalkou Str., 11581, Athens, Greece
vagelat@cti.gr

**Abstract**

The aim of this paper is to present the R&D activities carried out at Neurosoft S.A. regarding the development of proofing tools for Modern Greek. Firstly, we focus on infrastructure issues that we faced during our initial steps. Subsequently, we describe the most important insights of three proofing tools developed by Neurosoft, i.e. the spelling checker, the hyphenator and the thesaurus, outlining their efficiencies and inefficiencies. Finally, we discuss some improvement ideas and give our future directions.

## 1. INTRODUCTION

The evolution of Human Language Technology (HLT) is based on R&D activities in the following areas:

1. Language Resources: computational realizations of models that represent the components of human language (i.e. phonemes, syllables, morphemes, words, phrases, sentences, discourse) and the levels of language analysis (i.e. phonology, prosody, morphology, syntax, semantics, pragmatics), in the forms of lexicons, wordnets, ontologies, knowledge bases, text and speech corpora, rules, n-grams, decision trees, connectionist networks, etc.
2. Language Tools. They can be grouped into:
    a) Infrastructure Tools: software systems for the development of language resources, e.g. lexicographical databases, corpus management systems, rule-writing or machine-learning workbenches, etc.
    b) Application Tools: software components or systems built on top of language resources, utilized by end-users to satisfy information needs, e.g. lexicon browsers, search engines, or to perform automatic text or speech processing, e.g. text-to-speech converters, spelling/grammar/style checkers, summarizers, machine translators, etc.

Evidently, there is a long distance to cover till the delivery of any HLT application tool to end-users; the higher the language analysis level to be reached by the tool, the longer the distance. Furthermore, there are user requirements (e.g. for computers capable of "understanding" the human language) that cannot be effectively fulfilled at this moment, not even in the next decade, due to technology gaps.

Proofing tools are HLT application tools that help humans (typists, typesetters, writers, authors, translators, editors, etc.) to write, typeset, proofread, search, summarize and/or translate texts. They are incorporated into contemporary word processors or desktop publishing systems and are strictly language-specific, i.e. there is a version of each tool for each supported natural language. The complete suite of proofing tools for a specific natural language comprises:

- An electronic *dictionary/thesaurus* that provides word meanings, example uses, synonyms, antonyms, word translations, etc.
- A *hyphenator* that automatically syllabifies the end-of-line words, so as to avoid losing printable space during paragraph alignment.
- A *stemmer* that produces all morphological variations of a specific word. It is utilized for word search & replace functions and for query expansion during document retrieval.
- A *spelling checker* that locates words with orthographical errors in texts and suggests corrections.
- A *grammar checker* that locates ungrammatical constructions in texts and suggests corrections.
- A *style checker* that flags violations of style rules (e.g. when the text contains informal words or constructions with low readability) and suggests style alterations.
- A *summarizer* that produces the summary of a given text.
- A *translator* that translates a text from its original language to another language.

## 2. BACKGROUND

Since 1999, our HLT team at Neurosoft S.A. has worked in developing language resources and tools for Modern Greek (M. Greek). In these 4.5 years, we gave major emphasis –and effort– to infrastructure issues, i.e. to language resources and tools for developing these resources. The kick-off activity was to model: a) the graphemic components of M. Greek up to the word level (alphabet, syllables, morphemes, words), b) the type (phonetic, morphological, syntactic, semantic) and the (simple or complex) structure of information that can be assigned to each component and c) the inflectional and derivational system of M. Greek. The next step was to implement the above models and integrate them into a lexicographical database, using XML as the core description and content structuring language.

Towards the development of a morphological lexicon and a thesaurus for M. Greek, we extended the lexicographical infrastructure with a) a corpus of M. Greek texts (~100 million words) and b) a set of light-weight lemma-encoding GUIs. The corpus was initially used to count word frequencies and later to retrieve examples of words in context. The lemma-encoding GUIs allowed the lexicographers to work off-line (e.g. at home) and produce XML files ready to be uploaded in the central database.

The development of the morphological lexicon was divided into two phases: a) selection of the vocabulary, by excerpting the 4 major M. Greek dictionaries (Κριαρά [12], Μείζον Τεγόπουλου-Φυτράκη [17], Μπαμπινιώτη [14] and Λεξικό Α.Π.Θ. [13]) and counting word frequencies in the corpus, and b) production of all morphological forms for each word in the vocabulary. Inevitably, in both phases, we faced the perennial problem of language standardization, i.e. common M. Greek (standard) vs. katharevousa and dialects. On one hand, the dictionaries brought up many discrepancies about whether or not a word/word-form belongs to the standard language. On the other hand, the corpus revealed many words/word-forms that either were not included in any of the 4 dictionaries or were flagged by the dictionaries as divergences from the standard language. To date, after more than 120 person months, the morphological lexicon contains ~90,000 words (~1,100,000 word-forms) with orthographical, syllabification, morphological, morphosyntactic and morphostylistic information.

Having a lexical database with all the above information, the natural follow-up phase is to enrich it with semantic information. Developing a thesaurus is an important step forwards the difficult field of semantics; synonymy, antonymy, hyperonymy, hyponymy and meronymy are some very significant relations between concepts, and it is very useful (to humans and to computers) to have these relations recorded somewhere. The development of the thesaurus was divided into three phases: a) selection of lemmas that contain at least one synonym or antonym, b) distinction of the meanings of each lemma, and c) definition of synonyms, antonyms and example-uses (where needed) per meaning. All three phases were based on the excerption of the 4 major M. Greek dictionaries and on the extraction of word-in-context lines from the corpus and from the Internet. To date, after more than 40 person months, the thesaurus contains ~22,500 lemmas with synonyms, antonyms and example-uses per meaning. An interesting property of the thesaurus is its closure: there is always a lemma for any word that participates in a synonymic or antonymic relation.

Apart from being prerequisites for the advancement of HLT at Neurosoft, the aforementioned language resources provided the passport for entering the frontiers of proofing tools technology. After the morphological lexicon reached a satisfactory content level, we could immediately proceed to the development of at least 3 proofing tools for M. Greek, namely the spelling checker, the hyphenator and the stemmer. The spelling checker and the hyphenator have already become available for the following platforms: a) MS Office 97, 2000, XP / 98, X (MS Win / Mac OS), b) Sun Open Office, c) Adobe InDesign, Photoshop, Illustrator (MS Win / Mac OS) and d) Quark Xpress (MS Win / Mac OS). The forthcoming M. Greek thesaurus, which incorporates the functionality of the stemmer, will be initially available for MS Office and Sun Open Office, as well as a standalone tool with its proprietary browser.

## 3. MDAGs and TRIEs

The Minimal Directed Acyclic Graphs (MDAGs, [2], [6]) and the TRIEs ([3], [1]) are two variations of Finite State Automata (FSA, [2]) that have been thoroughly used in HLT as lexical representation structures. Their major characteristics, which substantiate their ability to store and manipulate large word sets, are:

- *Speed.* The speed of the lookup function depends on the length of the searched word and not on the size of the lexicon.
- *Sorting Convenience.* The words stored in an FSA can be easily sorted, by sorting the outgoing transitions of each node.
- *Regular Expression Support.* An FSA can easily evaluate complex regular expressions. This also permits the development of smart word correction algorithms, which utilize regular expressions to produce alternative words.

Both MDAG and TRIE represent common prefix paths. MDAG also represents common suffix paths, resulting to smaller automata (fewer states and transitions). Figure 1 illustrates the MDAG (a) and the TRIE (b) representations of six words (ισομετρία, ισομετρίας, ισομετρίες, ισομοιρία, ισομοιρίας, ισομοιρίες).

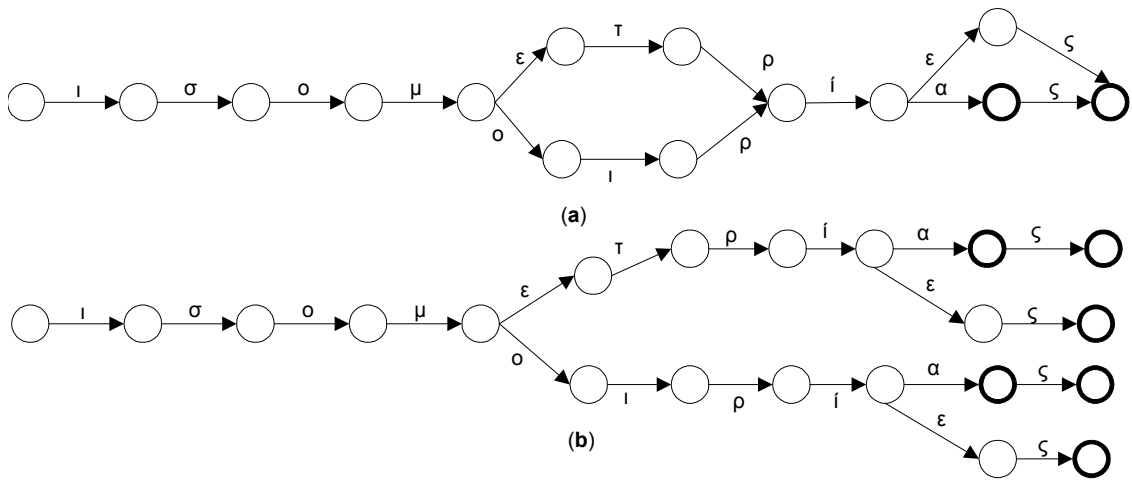

Figure 1. (a) MDAG and (b) TRIE

As shown in Figure 1, the MDAG structure (14 nodes, 15 transitions) is smaller than the TRIE structure (21 nodes, 20 transitions). Also, the MDAG has only two terminal nodes (the bold nodes) for the six words, whereas the TRIE has six terminal nodes, one for each word. This practically means that MDAG is more efficient for word storage, but TRIE can also be used as a record indexing structure: since each terminal node in the TRIE corresponds to a discrete word (this is false for MDAC), the TRIE is suitable for storing record key-words; what is further needed is some extra space in each terminal node to store a record pointer. We are using MDAG to store the words of our spelling lexicon and TRIE to index the lemmas of the thesaurus. More than 1.000.000 M. Greek word-forms (12Mb) were converted to a MDAC structure (790Kb), following a method similar to that of Mihov [7]. The search speed of this structure is ~600,000 words/second on a 1.7GHz MS Windows 2000 computer. The speed of the thesaurus index is similar but the size is much bigger. For 26,000 thesaurus lemmas, using ~400,000 word-forms as indexing keys, the size of the TRIE index is 5.6Mb.

## 4. SPELLING CHECKER

The spelling checker ascertains the orthographical correctness of an input word, if this word is found in its spelling lexicon. If not, the spelling checker has to produce a list of alternatives that are graphically or phonetically "similar" to the input word. This list is ordered according to a similarity degree; the alternatives that are most similar to the input word appear at the head of the list. Word similarity is calculated by a distance function. To stress the importance of the distance function, we claim that an optimal distance function on its own would be sufficient to carry out the entire correction process: given an incorrect word, we can calculate the distance between it and each word of the spelling lexicon and (according to a threshold) select the nearest words as alternatives.

The distance function we use measures the *Levenshtein distance* [5] (or *edit distance*) between two words, i.e. the number of deletions, insertions, or substitutions required to transform one word into the other. Véronis [11] proposes a different method, in which all the phonetically equivalent graphemes are equidistant, independently of the number of characters they differ. Another approach is that of Ristad and Yanilos [10], who define a stochastic model capable to learn the distance function from training examples. This reduces the error rate to one fifth of the corresponding Levenshtein error rate.

To calculate the distance of an incorrect word to all words of the spelling lexicon is very costly and thus inefficient. It would be convenient to select an appropriate subset of the lexicon first. What the spelling correction algorithms actually do is to alter the incorrect word (by inserting, deleting, substituting characters) and produce a set of strings; the intersection of this set with the entire lexicon provides the desirable lexicon subset, which passes through the distance function and gives a list of ordered alternatives. The alterations of the incorrect word performed by the correction algorithms are not arbitrary; they are based on the reasoning about spelling errors. In general, according to their causes, spelling errors fall in the following categories:

- *typographic*: the user, due to haste or even carelessness, types a wrong character or an extra character, misses a character or transposes two characters.
- *morphological*: the user does not know the morphology of the word he types.
- *pronunciation*: the user does not know the pronunciation of the word he types.
- *grammatical*: the user types an orthographically correct word, which is syntactically or semantically unrelated to the context.

The difficulty level of correcting a spelling error is analogous to the order of the above error categories. There are also other categories, such as human or machine optical recognition errors, data transmission errors, etc. These categories are considered domain or source specific and require special handling. Leaving for the future the category of *grammatical* errors, as it requires syntactic and/or semantic analysis, we developed a variation of Véronis' phonographic correction method [11], in order to handle *pronunciation* and *morphological* errors. To apply this method, we first categorize all the different graphemes in equivalence classes, as follows:

- Double consonants and the corresponding single consonants: { "λλ", "λ" }, { "κκ", "κ" }, { "μμ", "μ" }, { "νν", "ν" }, { "ρρ", "ρ" }, { "σσ", "σ" }, { "ττ", "τ" } and { "ππ", "π" }.
- Combinations of consonants with identical or similar articulation: { "πσ", "ψ" }, { "κσ", "ξ" }, { "γγ", "γκ" }, { "τσ", "τζ"}, etc.
- Vowel digraphs, vowel combinations and single vowels or vowel-consonant digraphs with identical articulation: {"ε", "έ", "αι", "αί"}, {"ι", "í", "ϊ", "ΐ", "η", "ή", "υ", "ύ", "ϋ", "ΰ", "ει", "εί", "οι", "οί", "υι", "υί"}, {"αυ", "αβ", "αφ", "αύ", "άβ", "άφ"}, etc.
- Optically similar graphemes: { "β", "θ" }, { "υυ", "ω" }, { "Ο", "Θ" }, { "Μ", "ΛΛ" }, etc.

All graphemes of the equivalence classes are used to construct a FSA, which will guide the creation of regular expressions for the production of word alternatives. An example of such FSA is illustrated in Figure 2:

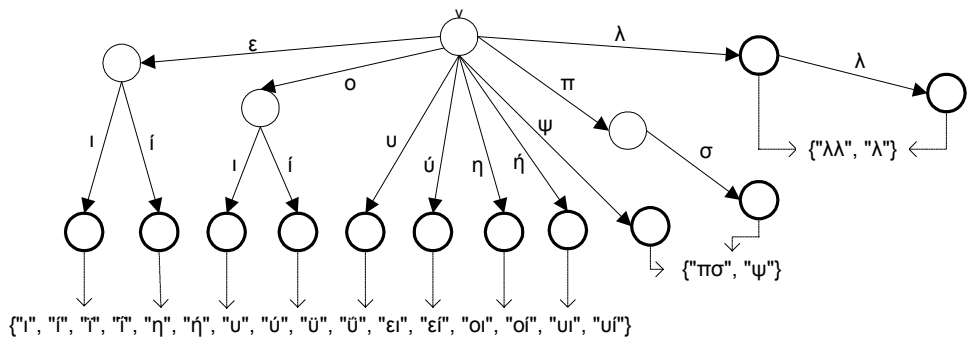

Figure 2. FSA for the formation of regular expressions

In case of an unknown input word, we use the above FSA to identify constituent graphemes that belong to equivalence classes. Each identified grapheme is substituted by all the graphemes of the equivalence class it belongs to, constructing this way a regular expression that will produce alternative words. For example, given the unknown word πσιχυ the above FSA will recognize the graphemes [πσ][ι]χ[υ]. By substituting the identified graphemes with their equivalence classes, we take the regular expression:

(πσ | ψ)(ι | ί | ϊ | ΐ | η | ή | υ | ύ | ϋ | ΰ | ει | εί | οι | οί | υι | υί)χ(ι | ί | ϊ | ΐ | η | ή | υ | ύ | ϋ | ΰ | ει | εί | οι | οί | υι | υί)

When this expression is searched in the spelling lexicon (MDAG), it produces the following list of valid M. Greek words (sorted according to the Levenshtein distance):

```
Incorrect word: πσιχυ    Suggested words: ψυχή (4), ψύχει (5), ψυχοί (5)
```

Algorithms that find alternatives for single typographic errors are simple and efficiently implemented. They produce a list of candidate alternatives by substituting letters of the unknown word and then use the lexicon to filter out the non-word candidates. A mixture of typographic, pronunciation and morphological errors is more difficult to be handled efficiently. A simple method is to run the algorithms that handle the typographic errors first and then pass all the produced candidates directly (without having been filtered by the lexicon) to the algorithms that handle the pronunciation/morphological errors.

Despite the fact that the combined application of all correction algorithms on an unknown word can produce extremely complex regular expressions, the list of alternatives (for any unknown word) is produced in less than 100 milliseconds (on a 1.7GHz Windows 2000 computer), thanks to the search speed of the MDAG structure (few milliseconds per regular expression).

## 5. HYPHENATOR

To develop a rule-based hyphenator for M. Greek seems to be a quite straightforward task, as M. Greek grammar [15] provides a set of syllabification rules that can be easily transformed into computer code. These rules, slightly modified to look like pseudo-code, are:

1) A syllable must contain at least one vowel[1].

2) A <vowel$_1$><consonant><vowel$_2$> sequence splits into <vowel$_1$> - <consonant><vowel$_2$> (e.g. θέ-**λ**ω, πα-**ρ**ά-**θ**υ-**ρ**ο).

3) A <vowel$_1$><consonant$_1$><consonant$_2$><zero_or_more_consonants><vowel$_2$> sequence splits into:
   a) <vowel$_1$> - <consonant$_1$><consonant$_2$><zero_or_more_consonants><vowel$_2$>, if there exists a M. Greek word that starts with <consonant$_1$><consonant$_2$> (e.g. α-**στρ**α-πή). The consonant bigrams that can be found at the beginning of M. Greek words are: βγ, βδ, βλ, βρ, γδ, γκ, γλ, γν, γρ, δρ, θλ, θν, θρ, κβ (e.g. **κβ**άντο), κλ, κν, κρ, κτ, μν, μπ, ντ, πλ, πν, πρ, πτ, σβ, σγ, σθ, σκ, σλ (e.g. **σλ**αβικός), σμ, σν (e.g. **σν**ομπάρω), σπ, στ, σφ, σχ, τζ, τμ, τρ, τσ, φθ, φλ, φρ, φτ, χθ, χλ, χν, χρ and χτ.
   b) <vowel$_1$><consonant$_1$> - <consonant$_2$><zero_or_more_consonants><vowel$_2$>, if no M. Greek word starts with <consonant$_1$><consonant$_2$> (e.g. εκ-**στρ**α-τεί-α).

4) A <vowel$_1$><vowel$_2$> sequence splits into <vowel$_1$> - <vowel$_2$>, if all the following are false:
   a) <vowel$_1$><vowel$_2$> is one of the following vowel combinations: αυ (/av/ or /af/), αύ (/áv/ or /áf/), ευ (/ev/ or /ef/), εύ (/év/ or /éf/), ηυ (/iv/ or /if/) or ηύ (/ív/ or /íf/).
   b) <vowel$_1$><vowel$_2$> is one of the following vowel digraphs, which are articulated as monophthongs[2]: αι (/ε/), αί (/έ/), ει (/i/), εί (/í/), οι (/i/), οί (/í/), υι (/i/), υί (/í/), ου (/u/) or ού (/ú/). The bigrams υι and υί constitute digraphs (e.g. **υι**-ός, κα-θε-στη-κ**υί**-α) only when they are not preceded by ο or ε; otherwise, the υ of υι or υί is combined with the preceding ο to form the digraph ου (/u/, e.g. ιν-δ**ου**-ι-σμός) or with the preceding ε to form the combination ευ (/εv/, e.g. Λ**ευ**-ί-της).
   c) <vowel$_1$><vowel$_2$> form a diphthong[3] or are part of a diphthong (e.g. ιο is a diphthong in ί-δ**ιο**ς but not in αιφ-νί-δι-**ο**ς, οι is part of a diphthong in ό-π**οι**οι but not in ό-μ**οι**-οι).

A bigram lookup table and a few if-then-else statements would have been sufficient to express the above rules in computer words, if we could disregard 4c. Strictly dependent on the phenomenon of synizesis, diphthongs are unambiguously recognizable only during oral communication (phonemic level): a vowel sequence is a diphthong if it is articulated as a single phoneme. But, in a written text (graphemic level), the only information available is the sequence of alphabetic characters that constitute each word. At the graphemic level, in order to decide whether a sequence of vowels form a diphthong or not, a hyphenator has no alternative but to examine the surrounding alphabetic characters.

---

[1] The Greek vowels are: α, ά, ε, έ, η, ή, ι, ί, ϊ, ΐ, ο, ό, υ, ύ, ϋ, ΰ, ω and ώ. The Greek consonants are: β, γ, δ, ζ, θ, κ, λ, μ, ν, ξ, π, ρ, σ, ς, τ, φ, χ and ψ.

[2] A *monophthong* is a "pure" vowel sound, one whose articulation at both beginning and end is relatively fixed, and which does not glide up or down towards a new position of articulation. (Taken from: http://en.wikipedia.org)

[3] A *diphthong* is a vowel combination usually involving a quick but smooth movement from one vowel to another, often interpreted by listeners as a single vowel sound or phoneme. (Taken from: http://en.wikipedia.org)

The identification of diphthongs during the computational syllabification of M. Greek words is not carried out effectively by existing approaches. In the hyphenation patterns prepared (manually) by Φιλίππου [18] for TEX and LATEX typesetting systems, a conservative policy is followed: instead of splitting certain vowel sequences that may (or may not) form a diphthong, better don't split them at all. In this way, non-diphthongs are considered diphthongs. The article of Noussia [8], about the rule-based hyphenator of MS Office, includes an extensive reference to the phenomenon of diphthong ambiguity and presents a set of 12 handcrafted rules for vowel hyphenation, which hyphenate correctly more than 89.9% of the vowel sequences that can possibly occur in M. Greek words. The exact degree of diphthong ambiguity resolution is not reported in [8], mainly because the overall diphthong ambiguity could not be measured (due to lack of an exhaustive list of hyphenated words).

It became apparent that, in order to cope with diphthong ambiguity on the whole, we should record/measure it first. Our study [16] was based on a set of 878,272 hyphenated word-forms, taken from our morphological lexicon (each word-form in this lexicon is accompanied by syllabification information). The longest vowel sequence (7 vowels!) was found in the loan word Τσι**ουάουα** (Chihuahua - Mexican dog breed). Table 1 illustrates all the sequences of 2-7 consecutive vowels that were found in the morphological lexicon[4]:

| (a) | (b) Valid Vowel Sequences | (c) | (d) |
|---|---|---|---|
| 2 | άα, άε, **άη**, **άι**, άο, άυ, άω, έα, έε, έη, έι, έο, έω, ήε, ήι, ίη, ήω, ία, ίε, ίη, ίο, ίυ, ίω, αΐ, αά, αέ, αή, αί, αϋ, αα, αε, **αη**, αι, αο, αυ, αω, **αϊ**, αϋ, αό, αύ, αώ, εΐ, εά, εέ, εή, εί, εϋ, εα, εε, εη, ει, εο, ευ, εω, **εϊ**, εϋ, εό, εύ, εώ, ηέ, ηί, ηε, ηο, ηυ, ηω, ηύ, **ηώ**, **ιά**, **ιέ**, ιή, ιί, **ια**, **ιε**, ιη, ιι, **ιο**, ιυ, **ιω**, **ιό**, ιύ, **ιώ**, οΐ, οά, οέ, οή, οί, οϋ, οα, οε, οη, οι, οο, ου, οω, **οϊ**, οϋ, οό, οΰ, οώ, υΐ, υά, **υέ**, υή, **υα**, υε, υη, **υο**, υω, υϊ, **υό**, υύ, **υώ**, ωά, ωέ, ωή, ωί, ωα, ωε, ωη, ωι, ωο, ωυ, ωω, ωό, ωύ, ωώ, όα, όε, **όη**, **όι**, όο, όυ, όω, ύα, ύε, ύη, ύι, ύο, ύω, ώα, ώε, ώη, ώι, ώο, ώω | 144 | 23 |
| 3 | άει, άια, άιε, άιο, άοι, άου, έαι, έει, έια, έιο, έιω, έοι, έου, ήια, ήιε, ήιη, ήιο, ήιω, ίαι, ίαυ, ίει, ίευ, ίοι, ίου, αΐα, αΐο, αΐω, αία, αίί, αίο, αίω, αεί, αει, αιά, αιέ, αιή, αιί, αια, αιε, αιη, αιι, αιο, αιω, αιό, αιώ, αοΐ, αοί, αοι, αου, αοΐ, αού, αυά, αυα, αυε, αυό, αϊα, αϊε, αϊο, αϊό, αύα, αύε, αύο, αύω, εΐι, εΐε, εΐω, εέα, εέω, είε, είο, είω, εαε, εαι, εαυ, εεί, εει, **ειά**, **ειέ**, ειή, ειί, **εια**, **ειε**, ειη, ει, **ειο**, **ειω**, **ειό**, **ειώ**, εοέ, εοί, εοα, εοε, εοι, εοο, εου, εοϊ, εοΰ, ευά, ευέ, ευή, ευί, ευα, ευε, ευη, ευι, ευο, ευυ, ευω, ευό, ευώ, εϊα, εϊο, εϊό, εϊώ, εύα, εύε, εύη, εύο, εύω, ηού, **ιάε**, **ιάο**, **ιέα**, **ιέω**, **ιαί**, **ιαε**, **ιαι**, **ιαο**, **ιαυ**, **ιαό**, **ιαύ**, **ιεί**, **ιει**, **ιεο**, **ιευ**, **ιεύ**, ιηύ, **ιιέ**, **ιια**, **ιιό**, **ιιώ**, **ιοί**, **ιοα**, **ιοε**, **ιοη**, **ιοι**, **ιου**, **ιοϊ**, **ιοϋ**, **ιού**, οΐα, οΐε, οέα, οέω, οία, οίε, οίη, οίο, οίω, οαί, οαε, οαι, οαυ, οαύ, οεί, οει, οεο, οευ, οεό, **οιά**, **οιέ**, οιή, **οια**, **οιε**, **οιη**, οιι, **οιο**, **οιω**, **οιό**, **οιώ**, οοί, οοι, ουα, ουε, ουί, ουα, ουε, ουη, ουι, ουο, ουυ, ουω, ουό, ουώ, οϊέ, οϊα, οϊω, οϊό, οϊώ, οϋά, οϋα, ούα, ούε, ούη, ούι, ούο, ούω, υΐα, υΐε, υαι, υαυ, υεί, υει, υευ, υεύ, υιά, υιέ, υια, υιε, υιι, υιο, υιω, υιό, υιώ, υοέ, **υοί**, υοε, υοι, υου, **υού**, υϊά, υϊα, υϊώ, υόε, ωία, ωίε, ωαι, ωεί, ωιώ, ωοί, ωοι, ωου, ωού, όαι, **όει**, όοι, όου, ύαι, ύαυ, ύει, ύευ, ύοι, ύου, ώει, **ώια**, **ώιε**, **ώιο**, **ώιω**, ώοι, ώου | 268 | 52 |
| 4 | άιοι, άουα, έιου, ήιοι, ήιου, ίαια, ίαιε, ίαιη, ίαιο, ίαιω, ίευα, ίευε, αΐου, αίει, αίευ, αίοι, αίου, αεία, αείο, αείω, αευα, αιευ, αιεύ, αιοί, αιοα, αιοε, αιοη, αιοι, αιου, αιοϋ, αιού, αύει, εΐαι, εΐει, εΐοι, εΐου, ειαί, ειαι, ειεί, **ειευ**, **ειεύ**, **εια**, εισε, ειος, ειοη, **ειοι**, **ειου**, **ειού**, εοει, εοου, εοϋο, ευάι, ευαί, ευαι, ευει, ευοι, ευου, ευοό, εωει, εωοι, εύει, εύου, **ιάου**, ιαία, ιαίε, ιαίο, ιαίω, ιαιο, ιαιω, ιαιό, ιαιού, **ιαου**, **ιαού**, ιεία, ιείε, ιείο, ιείω, ιειο, ιειώ, **ιευα**, **ιευε**, ιευο, ιευό, **ιεύα**, **ιεύε**, **ιεύο**, **ιεύω**, ιιοί, ιιού, ιοαέ, ιοαε, **ιοει**, **ιοευ**, **ιουά**, **ιοϊα**, όίοι, οίου, οαιώ, οεία, οείε, οείο, οείω, οειώ, οευε, οϊα, οϊέ, οία, οίε, οιαύ, οιεί, οιιώ, οιοί, **οιοι**, **οιου**, **οιού**, οοει, οοιω, ουάη, ουαί, ουοι, ουου, οϋϊ, οϋου, οϋοϋ, ούει, **ούια**, ούου, υαία, υαίε, υαίο, υαίοι, υιοί, υιοι, υιου, υιοΰ, υοει, υοϊε, υόει, ωοει, όεια, όειε, όειο, όειω, ώιοι, ώιου | 146 | 22 |
| 5 | ίαιοι, ίαιου, άίευα, άίευε, αείου, αιευό, αιευά, αιευέ, αιευό, αιευώ, αιοει, είευα, είευε, **ειευα**, **ειευε**, **ειευό**, **ειεύα**, **ειεύε**, **ειεύο**, **ειεύω**, ευοίω, ιαίοι, ιαίου, ιείοι, ιείου, **ιεύει**, **ιεύου**, οείου, οευαί, οευαι, οιεία, οιεία, οιείε, οιείο, οιειώ, οιειω, οιοει, ουάου, υαίοι, υαίου, όειοι, όειου | 41 | 9 |
| 6 | αιεύει, αιεύου, ειεύει, ειεύου, ιοευαι, οιείου | 6 | 0 |
| 7 | ιουάουα | 1 | 0 |
| | **Total:** | **606** | **106** |

Table 1. Sequences of consecutive vowels found in M. Greek words
     column (a): sequence length
     column (c): number of sequences
     column (d): number of sequences with diphthong ambiguity

The bold vowel sequences/sub-sequences of Table 1 are candidate diphthongs; in some words they do not split but in other words they do split. Independently of how many consecutive vowels occur in a word, according to syllabification rule 4 it is sufficient to decide whether <u>two</u> adjacent vowels split or not. For example, the only possible split of ειου is ει-ου (since neither ει nor ου split, due to rule 4b); the decision whether to insert a hyphen between ει and ου (i.e. whether ειου is a diphthong, as in ά-δ**ειου**, or not, as in ε-πι-τή-δ**ει**-**ου**) is computationally equivalent to the decision whether to insert a hyphen between ι and ο. Table 2 illustrates all the vowel bigrams (total 24) that were located in the list of 878,272 hyphenated word-forms, which exhibit syllabification ambiguity; all these bigrams either form diphthongs or are part of diphthongs.

---

[4] Words unknown to the morphological lexicon may contain vowel sequences not recorded in Table 1. Due to the large coverage of the morphological lexicon, we expect (without being able to measure this expectation) that such unknown vowel sequences will be very few.

|   | Vowel Bigram (a) | # (b) | Diphthongs that contain the Bigram (c) | Examples of non-Splitting (d) | % (e) | Examples of Splitting (f) | % (g) |
|---|---|---|---|---|---|---|---|
| 1 | ια | 49,650 | ια, εια, ιαί, ιαι, οια | φτώ-χεια, δό-λια, δια-χέ-ω | 44 | Α-ντι-ό-χει-α, δό-λι-α, δι-α-χέ-ω | 56 |
| 2 | ιο | 30,425 | ιο, ιοι, ιου, ιού, ιοί, ειο, οιο, ειοι, ειου, ειού, οιοι, οιου, οιού | ο-λό-ι-διος, κα-θά-ριοι, θειου | 19 | αιφ-νί-δι-ος, Ά-ρει-οι, υ-δρό-θει-ου | 81 |
| 3 | ιά | 25,784 | ιά, ειά, οιά | πιά-σω, βιά-ζουν, λιά-σω | 75 | κο-πι-ά-σω, βι-ά-ζουν, σχο-λι-ά-σω | 25 |
| 4 | ιώ | 13,649 | ιώ, ειώ, οιώ | λο-γιών, Λη-ξου-ριώ-της | 31 | τε-χνο-λο-γι-ών, Λαυ-ρι-ώ-της | 69 |
| 5 | ιε | 9,302 | ιε, ειε, ιεί, ιει, οιε | άγιε, γιε, ή-πιε, ό-ποιες | 11 | ά-γι-ε, ή-πι-ε, ό-μοι-ες | 89 |
| 6 | ιω | 9,266 | ιω, ειω, οιω | ί-σιω-να, τε-λειω-μέ-νος | 18 | α-παί-σι-ων, τε-λει-ω-μέ-νος | 82 |
| 7 | ιό | 8,241 | ιό, ειό, οιό | κα-τα-ριό-ταν, θε-ριό | 57 | κα-θα-ρι-ό-τη-τα, θη-ρι-ό-μορ-φος | 43 |
| 8 | υό | 7,584 | υό | δυό-μι-σι, κα-ρυό-φυλ-λο | 0,6 | α-να-δυ-ό-μα-στε, κα-ρυ-ό-τυ-πος | 99,4 |
| 9 | υα | 4,189 | υα | α-μυα-λιά, ξε-στά-χυα-σα | 14 | μυ-αλ-γί-α, πε-ρι-στά-χυ-α | 86 |
| 10 | ιέ | 4,168 | ιέ, ειέ, οιέ | αλ-λα-ξιές, πιέ-στε, θε-ριέ-ψω | 65 | δε-ξι-ές, πι-έ-στε, α-γρι-έ-ψω | 35 |
| 11 | αϊ | 2,751 | αϊ | αϊ-τός, χαϊ-μα-λί, νε-ραϊ-δί-σιος | 18 | σα-ϊ-τεύ-ω, α-χα-ϊ-κός, πα-ρα-ϊ-α-τρι-κός | 82 |
| 12 | υο | 1,420 | υο, υοί, υού | δυο-νών, κα-ρυο-φύλ-λι | 0,7 | α-σό-δυ-ο, κρυ-ο-φθο-ρι-σμός | 99,3 |
| 13 | υά | 1,178 | υά | μα-το-γυά-λια, φτυά-ρι | 15 | μα-νου-ά-λια, φλυ-ά-ρη-σα | 85 |
| 14 | οϊ | 1,052 | οϊ | βοϊ-δί-σιος, κο-ροϊ-δεύ-ω | 17 | ευ-βο-ϊ-κός, μι-κρο-ϊ-δι-ο-κτή-της | 83 |
| 15 | εϊ | 593 | εϊ | λεϊ-μο-νιά, ζεϊ-μπέ-κι-κος | 8 | πλε-ϊ-μέ-ι-κερ, κα-ζε-ϊ-νι-κός | 92 |
| 16 | αη | 485 | αη | αη-δό-νι, καη-μός | 51 | α-η-δί-α, δε-κα-η-μέ-ρου | 49 |
| 17 | όη | 433 | όη | βόη-θη-σα, κα-λόηρ-θα | 5 | βό-η-σα, κα-λό-η-χα, α-νό-η-τα | 95 |
| 18 | υώ | 343 | υώ | δι-χτυών, λε-πτο-κα-ρυών | 1 | δι-κτυ-ώ-νω, Κα-ρυ-ώ-ν | 99 |
| 19 | άι | 280 | άι | χάι-δε-μα, χρυ-σο-γάι-τα-νο | 22 | εξ-αρ-χά-ι-σα, κι-λο-μπά-ιτ | 78 |
| 20 | υέ | 270 | υέ | λε-πτο-κα-ρυές | 0,8 | ι-δι-ο-φυ-ές, μυ-έ-λι-νος, σου-έτ | 99,2 |
| 21 | όι | 156 | όι | βόι-δι, ρόι-δι, κο-ρόι-δο | 22 | κον-βό-ι, πο-λα-ρό-ιντ, ο-λό-ι-διο | 78 |
| 22 | άη | 92 | άη | κε-λάη-δη-μα, κε-λάη-δη-σα | 44 | Μά-η-δες, χα-ρα-μο-φά-η-δες | 56 |
| 23 | όε | 84 | όει | κα-λόει-δα, α-πόει-δα | 10 | ι-στι-ο-πλό-ε, πρό-ε-δρος, α-θρό-ες | 90 |
| 24 | ηώ | 13 | ηώ | κα-ληώ-ρα | 24 | η-ώ, προ-σνη-ώ-σε-ως | 76 |
| | Total: | 171,408 | of 878,272 (19.51%) | | 37 | | 63 |

Table 2. Vowel bigrams with syllabification ambiguity
column (b): number of bigram occurrences
column (e): percentage of non-splitting occurrences
column (g): percentage of splitting occurrences

The vowel bigrams of Table 2 are sorted on column (b); the most frequent bigram is **ια**. The last line of Table 2 says that 19.51% of the 878,272 hyphenated word-forms contain at least one ambiguous vowel bigram; by generalizing this measurement, a M. Greek word is likely to exhibit diphthong ambiguity with ~0.2 probability. Also, on average, an ambiguous bigram splits in 37% and does not split in 63% of its occurrences.

For each of the above 24 ambiguous vowel bigrams we built a decision tree [9], using the hyphenated word-forms in which the bigram appears as training patterns. We then developed a hybrid M. Greek hyphenator that combines the following models:

a. Handcrafted rules that correspond to the syllabification rules 1-4b.
b. Decision trees that resolve the diphthong ambiguity introduced by the syllabification rule 4c.
c. An exception list with ~2,700 hyphenated word-forms. All these word-forms contain ambiguous vowel bigrams, which:
   - are handled incorrectly by the decision trees, or
   - when split the meaning of the word-form changes (i.e. the ambiguous bigrams appear in heterophonic homographs), e.g. ά-δεια (permission) and ά-δει-α (empty), χρό-νια (years) and χρό-νι-α (chronic), ή-λιο (sun) and ή-λι-ο (helium), σκιά-ζω (frighten) and σκι-ά-ζω (shade). We follow the conservative approach and do not split the bigrams in such word-forms.

Our hyphenator syllabifies correctly all the word-forms of the morphological lexicon. Taking into account that the decision trees hyphenate incorrectly the ~2,700 words of the exception list, i.e. 0,3% of the 878,272 word forms, the expected average error rate of the hyphenator on words never seen before is at most 0,3%.

# 6. THESAURUS

It happens very often to try to express our thoughts and the appropriate words do not come to our minds; we use some words that make sense, but we feel that they do not fit in the context. Also, we frequently realize that we have written the same word five times in a single paragraph, for no special reason but because we could not recall quickly an alternative with the same meaning. The role of the thesaurus is to help its user to overcome the above problems: given a word, the thesaurus returns a list of meanings; each meaning comprises an ordered list of synonyms; the first synonym of the list is the best alternative for the specific meaning.

From a theoretical point of view, the linguists argue that very few words have real synonyms, in the sense that *a* and *b* are synonyms if we can use *a* instead of *b* or *b* instead of *a* in whatever context. Then, one of *a* or *b* is redundant and is doomed to disappear as language evolves through the ages. From a practical point of view, the synonyms that thesaurus provides are *contextual* synonyms: we can use *a* instead of *b* only in certain contexts. That is why in a thesaurus it is essential to distinguish the synonyms of each word according to its meanings. This way, a thesaurus can also be used as a minimal semasiological dictionary (it describes the meanings of a word with synonym lists), but with caution, as there do not exist synonyms for every meaning.

Apart from offering the functionality described above, the thesaurus that we developed for M. Greek has the following characteristics:

- Contains ~22,500 lemmas.
- Each lemma is represented by a headword, which is the canonical form[5] of the word/phrase the lemma is about, but is accessible through any morphological form of the word/phrase. This is feasible because all the morphological forms of a lemma are used as indexing terms.
- The headword is accompanied by stylistic and domain information, e.g. the verb αγκαζάρω is informal, the noun αιμοσφαιρίνη is a term of Biology.
- The meanings of a lemma also contain antonyms (where possible) and example uses (where needed).
- Any word that appears in the synonyms or antonyms has always a corresponding lemma, i.e. any synonym or antonym is also a lemma headword.
- Not all the morphological forms of a word carry the same meaning(s). For example, αγκυλώνω means a) τσιμπάω, κεντάω, τρυπάω, βελονιάζω and b) καθηλώνω, παραλύω, παγώνω; but αγκυλώνομαι, which represents the passive forms of αγκυλώνω, also means παθαίνω αγκύλωση, πιάνομαι, a meaning that cannot be assigned to the active forms of αγκυλώνω. Such cases are coded as separate lemmas, i.e. there is a separate lemma for αγκυλώνομαι, which also contains a related-word reference to αγκυλώνω.

# 7. DISCUSSION – FUTURE DIRECTIONS

In the spelling checker section, we highlighted the importance of the distance function. The outperforming distance function proposed by Ristad and Yanilos [10] requires an extensive list of {incorrect word, correct word} pairs, which are used as training patterns. We have already started to collect such training patterns, so as to fulfill two goals: a) develop a better distance function and b) study the spelling errors methodically and conclude to an in-depth reasoning about them. The impression we obtain from what we have collected up to now is that spelling errors are strongly related to the idiosyncrasy of the user who causes them. An interesting enhancement of the spelling checker is to become capable of learning the idiosyncrasy of its user. A method to achieve this is to develop a spelling checker that monitors what suggested alternatives are adopted by the user; as the spelling checker knows which correction algorithms produced the preferred alternatives, in the future it can give higher priority to alternatives produced by these algorithms (the priority can be a parameter of the distance function)

As far as the hyphenator is concerned, we think that its performance (99.7%) has reached an upper limit. Hyphenation errors can occur only in words that contain ambiguous vowel bigrams and are unknown to the morphological lexicon. As the enrichment of the morphological lexicon is a live process, after having added a significant amount of new words, we will re-train the decision trees that handle the ambiguous vowel bigrams with the enriched word sets extracted from the morphological lexicon.

Thesaurus offers very little space for functional (algorithmic) improvement, but very large space for content improvement. As happens with every lexicon, the contents of thesaurus need nonstop amendment and enhancement. Our future plans about thesaurus include: a) addition of new lemmas, b) addition of is-a and part-of relations between lemmas, c) systematic review of the synonymic and antonymic relations between lemmas and d) addition of more example uses where needed.

---

[5] That is the singular, nominative form for nouns, the 1st person, singular, present, indicative, active form for verbs

What normally comes next is the development of a grammar checker for M. Greek. As previously stated, there are spelling errors that are not handled yet, which belong to the category of the grammatical errors. There are also syntactic or semantic errors that are grammatical in their nature. We have already started studying the grammatical errors in running texts and designing algorithms to handle them. We hope to have a grammar checker prototype in the next year.